\pdfoutput=1

\documentclass{article}	

\usepackage{hyperref}
\usepackage{subfigure}
\usepackage{url}
\usepackage[utf8]{inputenc} 
\usepackage{geometry} 
\usepackage{amsmath, amssymb, graphicx,times,bbm,algorithm,algorithmic,color}

\newcommand{\cU}{\mathcal{U}}

\newcommand{\EE}{\mathbb{E}}

\newcommand{\PP}{\mathbb{P}}
\newcommand{\RR}{\mathbb{R}}
\newcommand{\indic}{{\bf 1}}

\newcommand{\bp}{\noindent{\textbf{Proof.}}\ }
\newcommand{\ep}{\hfill $\Box$}

\newcommand{\eq}[1]{ \begin{equation} #1  \end{equation}}
\newcommand{\als}[1]{ \begin{align*} #1  \end{align*}}
\newcommand{\eqs}[1]{ \begin{equation*} #1  \end{equation*}}
\newcommand{\sk}{\nonumber\\}

\newcommand{\el}{\end{flushleft}}
\newcommand{\bl}{\begin{flushleft}}

\newcommand{\figsize}{0.48\columnwidth}

\newtheorem{fact}{Fact}	
\newtheorem{proposition}{Proposition}
\newtheorem{theorem}{Theorem}
\newtheorem{lemma}{Lemma}
\newtheorem{corollary}{Corollary}

\newtheorem{assumption}{Assumption}

\begin{document}
\title{A Streaming Algorithm for Crowdsourced Data Classification}
\author{Thomas Bonald $^\dag$ and Richard Combes $^\times$ 
\thanks{$\dag$: Telecom ParisTech, Networking \& Computer Science Department, Paris (France), {thomas.bonald@telecom-paristech.fr}} 
\thanks{$\times$: Centrale-Supelec and L2S, Telecommunication Department, Gif-sur-Yvette (France), {richard.combes@centralesupelec.fr}}}
\renewcommand\footnotemark{}
\renewcommand\footnoterule{}

\maketitle

\begin{abstract}
We propose a streaming algorithm for the binary classification of data  based on crowdsourcing. The algorithm learns the competence of each labeller by comparing her labels to those of other labellers on the same  tasks and uses this information to minimize the prediction error  rate on each task.
We provide performance guarantees of our algorithm for a fixed population of independent labellers.
In particular, we show that our algorithm is  optimal in the sense that the cumulative regret compared to the optimal decision with known labeller error probabilities   is finite, independently of the number of tasks to label. 
The complexity of the algorithm is linear in the number of labellers and the number of tasks, up to some logarithmic factors. Numerical experiments   illustrate the performance of our algorithm compared to existing algorithms, including simple majority voting and expectation-maximization algorithms, on both synthetic and real datasets. 
\end{abstract}

{\bf Keywords:} Crowdsourcing; data classification; streaming algorithms; statistics; machine learning.

\section{Introduction}

The performance of most machine learning techniques, and in particular  data classification, strongly depends on the quality of the labeled data used in the initial training phase.
A common way to label new datasets is through crowdsourcing: many people are asked to label data, typically texts or images, in exchange of some low payment. Of course, crowdsourcing is prone to errors due to the difficulty of some classification tasks,  the low payment per task and the repetitive nature of the job. Some labellers may even introduce errors on purpose. Thus it is essential to assign the same classification task to several labellers and to learn the competence of each labeller through her past activity so as to minimize the overall error rate and  to improve the quality of the labeled dataset.

Learning the competence of each labeller is a tough problem because the true label of each task, the so-called ``ground-truth", is unknown (it is precisely the objective of crowdsourcing to guess the true label). Thus the 
competence of each labeller must be inferred from the comparison of her labels on some set of tasks  with those of other labellers on the same set of tasks.

In this paper, we  consider binary labels and propose a novel algorithm for learning the error probability of each labeller based on the correlations of the labels. Specifically, we infer the  error probabilities of the labellers from  their  {\it agreement rates}, that is for each labeller the proportion of other labellers whom agree with her.
A key feature of this agreement-based  algorithm is  its streaming nature: it is not necessary to store the labels of all tasks, which may be expensive for large datasets. Tasks  can be classified on the fly, which simplifies the implementation of the algorithm. The algorithm can also be   easily adapted to  non-stationary environments where the labeller error probabilities   evolve over time, due for instance to the self-improvement of the labellers or to changes in the type of  data to label.
The complexity of the algorithm is linear, up to some logarithmic factor. 

We provide performance guarantees of our algorithm for a fixed population of labellers, assuming each labeller works on each task with some fixed probability and provides the correct label with some other fixed, unknown probability, independently of the other labellers. 
 In particular, we show that our algorithm is optimal in terms of {\it cumulative regret}, namely  the number of labels that 
 are different from those given by the optimal decision, assuming the labeller error rates are perfectly known, is finite, independently of the number of tasks. 
We also propose a modification of the algorithm suitable for non-stationary environments and provide performance guarantees in this case as well.
Finally, we compare the performance of our algorithm to those of existing algorithms, including simple majority voting and expectation-maximization  algorithms, through numerical experiments using both synthetic and real datasets.

The rest of the paper is organized as follows. We present the related work in the next section. We then describe the model and the proposed algorithm. Section \ref{sec:perf} is devoted to the performance analysis and  Section \ref{sec:nonstat} to the adaptation of the algorithm to non-stationary environments. The numerical experiments  are presented in Section \ref{sec:num}. Section \ref{sec:conc} concludes the paper.

\section{Related Work}

The first problems of data classification using independent labellers  appeared in the medical context, where each label  refers to the state of a patient (e.g., sick or sane) and the labellers are  clinicians.
In \cite{DS79}, Dawid and Skene proposed an expectation-maximization (EM) algorithm, admitting that the accuracy of the estimate was unknown. Several versions and extensions of this algorithm have since been proposed and tested in various settings \cite{HW1980,smyth1995,AD2004,raykar2010,LPI12}, without any significant progress on the theoretical side.
Performance guarantees have been  provided only recently  for an improved version of the algorithm relying on spectral methods in the initialization phase  \cite{ZCZJ14}.

A number of  Bayesian techniques have also been proposed and applied to this problem, see \cite{raykar2010,welinder2010,KOS11,LPI12,KOS13,KOS14} and references therein.
Of particular interest is the belief-propagation (BP) algorithm of Karger, Oh and Shah  \cite{KOS11}, which is provably order-optimal in terms of the number of labellers required per task for any given target error rate, in the limit of an infinite number of tasks and an infinite  population of labellers. 
 
Another family of algorithms is based on the spectral analysis of some matrix representing the correlations between tasks or labellers.
Gosh, Kale and McAfee \cite{GKM11} work on the task-task matrix whose entries correspond to the number of labellers having labeled  two tasks in the same manner, while Dalvi et~al.~\cite{DDKR13} work on the labeller-labeller matrix whose entries correspond to  the number of tasks  labeled  in the same manner by two labellers. Both obtain performance guarantees by the perturbation analysis of the top eigenvector of the corresponding expected matrix. 
The BP  algorithm of 
Karger, Oh and Shah is in fact closely related to these spectral algorithms: their message-passing scheme is very similar to the power-iteration method applied to the 
task-labeller matrix, as observed in \cite{KOS11}.

A recent paper proposes an algorithm based on the notion of minimax conditional entropy \cite{ZL15}, based on some probabilistic  model jointly parameterized by the labeller ability and the task difficulty. The algorithm is evaluated  through numerical experiments on real datasets only;  no theoretical results are provided on   the  performance  and the  complexity of the algorithm. 

All these  algorithms require the storage of all labels in memory. To our knowledge, the only streaming algorithm that has been proposed for crowdsourced data classification is the  recursive EM algorithm of Wang et al.~\cite{WAKA13}, for which no performance guarantees are available.

Some authors  consider slightly different versions of our problem. 
Ho et~al.~\cite{ho2012,ho2013} assume that the ground truth is known for some tasks and use the corresponding data to learn the competence of  the labellers in the exploration phase and  to assign tasks optimally in the exploitation phase. Liu and Liu \cite{LL15} also look for the optimal task assignment but without the knowledge of any true label: an iterative algorithm similar to EM algorithms is used to infer the competence of each labeller, yielding a cumulative regret in $O(\log^2t)$ for $t$ tasks compared to the optimal decision. Finally, some authors seek to {rank} the labellers with respect to their error rates, an information which is useful for  task assignment  but not easy to exploit for data classification itself \cite{chen2013,parisi2014}.

\vspace{1cm}

\section{Model and Algorithm}
\subsection{Model}
Consider $n$ labellers, for some integer $n>2$. Each task consists in determining the answer to a binary question. The answer to task $t$, the ``ground-truth",  is denoted by $G(t) \in \{+1,-1\}$. We assume that the random variables $G(1), G(2),\ldots$ are i.i.d.~and centered,  so that  there is no bias towards one of the answers. 

Each labeller provides an answer with probability $\alpha \in (0,1]$. When labeller $i \in \{1,...,n\}$ provides an answer, this answer is incorrect with probability  $p_i \in [0,1]$, independently of other labellers: 
$p_i$ is the error rate of labeller $i$, with 
 $p_i=0$ if labeller $i$ is perfectly accurate, $p_i=\frac 12$ if labeller $i$ is non-informative and $p_i =1$ if labeller $i$ always gives the wrong answer.
We denote by $p$ the vector $(p_1,\ldots,p_n)$. 

Denote by $X_i(t) \in \{-1,0,1\}$ the output of labeller $i$ for task $t$, where the output $0$ corresponds to the absence of an answer. We have:
$$
X_i(t) = \begin{cases} G(t) &\text{ w.p. } \;\;\; \alpha(1-p_i), \\ -G(t) &\text{ w.p. } \;\;\; \alpha p_i, \\ 0 &\text{ w.p. } \;\;\; 1-\alpha. \end{cases}
$$
Since the labellers are independent, the random variables $X_1(t),...,X_n(t)$ are independent given $G(t)$, for each task $t$. 
We denote by $X(t)$ the corresponding vector.
The goal is to estimate the  ground-truth $G(t)$ as accurately as possible by designing an estimator ${\hat G}(t)$ that minimizes the error probability $\PP( {\hat G}(t) \neq G(t))$. The estimator ${\hat G}(t)$ is adaptive and may be a function of $X(1),...,X(t)$ and the parameter $\alpha$ (which is assumed known), but cannot depend on $p$ which is a latent parameter in our setting.

\subsection{Weighted majority vote}

It is well-known that, given $p$ and $\alpha=1$, an  optimal estimator of ${G}(t)$ is the weighted majority vote \cite{N82,SG84}, namely
\begin{equation}\label{eq:wmv}
  \hat{G}(t) = \indic \{  W(t) > 0 \}- \indic \{  W(t) < 0 \}+  Z  \indic \{  W(t) = 0 \},
\end{equation}
where $W(t)= \frac 1 n\sum_{i=1}^n w_i X_i(t)$, 
$w_i= \log( 1/{p_i} - 1)$ is  the weight of labeller $i$ (possibly infinite), and $Z$ is a Bernoulli random variable of parameter $\frac 12$ over $\{+1,-1\}$ (for random tie-breaking).
We provide a proof that accounts for the fact that labellers may not provide an answer for each task. 

\begin{proposition}\label{pr:weightedmajority}
Assuming $p$ is known,  the estimator \eqref{eq:wmv} is an optimal estimator of $G(t)$. 
\end{proposition}
\bp
Finding an optimal estimator of  ${G}(t)$ amounts to finding an optimal statistical test between hypotheses $\{G(t) = +1\}$ and $\{G(t) =-1\}$, under a symmetry constraint so that type I and type II error probability are equal. Consider a sample $X(t) = x \in \{-1,0,1\}^n$ and denote by $L^{+}(x)$ and $L^-(x)$ its likelihood under hypotheses $\{G(t) = +1\}$ and $\{G(t) = -1\}$, respectively. 
We have
\als{
L^+(x) &= \prod_{i=1}^n (\alpha p_i)^{\indic\{x_i = -1\}} (\alpha (1-p_i))^{\indic\{x_i = 1\}} (1-\alpha)^{\indic\{x_i = 0\}}, \sk
L^{-}(x) &= \prod_{i=1}^n  (\alpha p_i)^{\indic\{x_i = 1\}} (\alpha (1-p_i))^{\indic\{x_i = -1\}} (1-\alpha)^{\indic\{x_i = 0\}}. 
}
We deduce the log-likelihood ratio, $$\log\left(\frac {L^+(x) }{ L^{-}(x)}\right) = \sum_{i=1}^n w_ix_i= w^Tx.$$ By the Neyman-Pearson theorem, for any level of significance, there exists $a$ and $b$ such that the uniformly most powerful test for that level is:
$$
 \indic \{  w^Tx > a \}- \indic \{  w^Tx < a \}+  Z  \indic \{  w^Tx = a \},
$$
where   $Z$ is a Bernoulli random variable of parameter $b$ over $\{+1,-1\}$. By symmetry, we must have $a = 0$ and $b=\frac 12$, which is the announced result. 
\ep

This result shows that estimating the true answer $G(t)$ reduces to estimating the latent parameter $p$, which is the focus of the paper.

\subsection{Average error probability}

A critical parameter for the estimation of $p$  is the  average error probability,
$$
q= \frac 1 n \sum_{i=1}^n p_i.
$$
We assume the following throughout the paper:

\begin{assumption}\label{a:fixedpoint}
	 We have $q < \frac 1 2 -\frac 1 n$.
\end{assumption}

This assumption  is essential. First, it is necessary to assume that $q< \frac 12$, i.e.,  labellers say ``mostly the truth". Indeed, the transformation $p \mapsto 1 - 2p$ does not change the distribution of $X(t)$, meaning  that the parameters $p$ and $1 - 2p$ are  statistically indistinguishable: it is the assumption $q< \frac 12$ that breaks the symmetry of the problem and allows one to distinguish between true and false answers.

Next, the accurate estimation of $p$ requires that there is enough correlation between the labellers' answers. Taking $p=(0,\frac 12,\ldots,\frac 12)$ for instance, the mean error rate is $q=\frac 1 2 -\frac 1 {2n}$ but the estimation of $p$ is  impossible since any permutation of the indices of $p$ lets the distribution of $X(t)$ unchanged.
For $p=(0,0,\frac 12,\ldots,\frac 12)$, the average error probability becomes $q=\frac 1 2 -\frac 1 {n}$, the maximum value  allowed by Assumption \ref{a:fixedpoint}, and     the estimation becomes feasible.

\subsection{Prediction error rate}

Before moving to the estimation of $p$, we give upper bounds on the prediction error rate, that is the probability that   $\hat G(t)\ne G(t)$,  given some estimator $\hat p$ of $p$.

First consider the case  $\hat p=(\frac 12,\ldots,\frac 12)$, which is a natural choice when nothing is known about $p$. The corresponding weights $\hat w_1,\ldots,\hat w_n$ are then equal and the estimator $\hat G(t)$ boils down to majority voting.
We get
\begin{align*}
\PP( {\hat G}(t) \neq G(t)) \le \PP(\sum_{i=1}^n X_i(t)\le 0|G(t)=1)
\le \exp\left({-\frac n 2 (\alpha(1-2q))^2}\right),
\end{align*}
 the second inequality following from Hoeffding's inequality.
For any fixed $q<1/2$, the prediction error probability decreases exponentially fast with  $n$. 

Now let $\hat p\in (0,1)^n$. The corresponding weights $\hat w_1,\ldots,\hat w_n$ are finite and the estimate $\hat G(t)$  follows from weighted majority voting. Again,
\begin{align*}
\PP( {\hat G}(t) \neq G(t))\le \PP\left(\sum_{i=1}^n \hat w_iX_i(t)\le 0|G(t)=1\right) 
\le \exp\left({-\frac 1 2 \frac{(\alpha\sum_{i=1}^n \hat w_i(1-2p_i))^2}{\sum_{i=1}^n \hat w_i^2}}\right),
\end{align*}
 the second inequality following from Hoeffding's inequality.

Consider for instance the ``hammer-spammer" model where $\alpha=1$ and $p=(0,\ldots,0,\frac 12,\ldots,\frac 12)$, i.e.,  half of the labellers always tell the truth while the other half always provide random answers. We obtain upper bounds on the prediction error rate equal to $e^{-n/8}$ for $\hat p=(\frac 12,\ldots,\frac 12)$ and  $e^{-n/4}$ for $\hat p\to p$. Taking $n=20$ for instance, we obtain respective bounds on the prediction error rate equal to $e^{-2.5}\approx 0.08$ and $e^{-5}\approx 0.007$: assuming these bounds are tight, this means that the accurate estimation of $p$ may decrease the prediction error rate  by an order of magnitude. 

\subsection{Agreement-based algorithm}
\subsubsection*{Maximum likelihood}

We are interested in designing an estimator of $p$ which has low complexity and may be implemented in a streaming fashion. 
The most natural way of estimating $p$ would be to consider the true answers $G(1),...,G(t)$ as latent parameters, and to calculate the maximum likelihood estimate of $p$ given the observations $X(1),...,X(t)$. The likelihood of a sample $x(1),...,x(t)$ given $G(1)=g(1),\ldots,G(t)=g(t)$ is \eqs{
	\prod_{s=1}^t \left(L^{+}(x(s))\indic\{g(s)=+1\} + L^{-}(x(s)) \indic\{g(s)=-1\}\right).
}
This approach has  two drawbacks. First, there is no simple sufficient statistic, so that one must store  the whole sample $x(1),...,x(t)$, which incurs a memory space  of ${O}(nt)$ and prevents any implementation through a streaming algorithm. Second, the likelihood is expressed as a product of sums, so that the maximum likelihood estimator is hard to compute, and one must rely on iterative methods such as EM.

\subsubsection*{Agreement rates}

We propose instead to estimate $p$ through the vector $a$ of  {\it agreement rates}. We define the agreement rate  of  labeller $i$ as  the average proportion  of other labellers whom agree with $i$, i.e., 
\begin{align}
a_i&=\frac 1 {n-1}\sum_{j\ne i}\PP(X_i(t)X_j(t)=1|X_i(t)X_j(t)\ne 0),\nonumber \\
&=\frac 1 {n-1} \sum_{j\ne i} (p_ip_j+(1-p_i)(1-p_j)).\label{eq:ri}
\end{align}
Observe that $a_i\in [0,1]$, with $a_i=0$ if labeller $i$ never agrees with the other labellers and $a_i=1$ if labeller $i$ always agrees with the other labellers.

Using the average error probability $q$, we get
$$
a_i=\frac 1 {n-1}(p_i ( n q - p_i) +  (1-p_i)( n-1-nq + p_i)),
$$
so that 
\begin{equation}\label{eq:trinome}
2p_i^2 - 2 p_i(  n(q - \frac 12) + 1) +nq -(1- a_i)(n-1)=0.
\end{equation}
For any fixed  $a_i$ and $q$, we see that $p_i$ is a solution to a quadratic equation; in view of Assumption \ref{a:fixedpoint}, this is the unique non-negative solution to this equation.

\subsubsection*{Fixed-point equation}

For any $u\in [0,1]^n$ and $v\in \RR$, let
$$
\delta_i(u,v)=v + 4\frac {n-1}{n^{2}} ( 1-2  u_i).
$$
Observe that this is the discriminant of the quadratic equation \eqref{eq:trinome}  for $u=a$ and $v=(2q-1)^2$.  It is non-negative   whenever $v\ge v_0(u)$, with
$$
	v_0(u)= \max(4\frac {n-1}{n^{2}}\max_{i=1,\ldots,n}( 2  u_i -1),0). 
$$
 Define the function $f$ by 
	\begin{align*}
\forall u, \forall v\ge v_0(u),\quad 	f(u,v) = \left(\frac 1 {n-2} \sum_{i=1}^n  \sqrt{\delta_i(u,v)} \right)^{2}.
		\end{align*}

\begin{proposition}\label{prop:fps}
The mapping $v \mapsto f(u,v) - v$ is strictly increasing over $[v_0(u),+\infty)$.
\end{proposition}
\bp
For any $u \in [0,1]^n$ and  $v>v_0(u)$, we have $\delta_i(u,v) > 0$ for all $i$, so that $v \mapsto f(u,v)$ is differentiable and its partial derivative is:
\als{
\frac{\partial f}{\partial v} (u,v)&= \frac{1}{(n-2)^2} \left( \sum_{i=1}^n  \sqrt{\delta_i(u,v)} \right) \left( \sum_{i=1}^n  \frac 1{\sqrt{\delta_i(u,v)}} \right).
}
Using Fact \ref{harm}, we obtain
$$
\frac{\partial f}{\partial v}   (u,v) \geq \frac{n^2}{(n-2)^2}>1.
$$
\ep 

\begin{fact}\label{harm}
For any positive real numbers $\chi_1,\ldots,\chi_n$,
$$
 \left( \sum_{i=1}^n  {\chi_i} \right) \left( \sum_{i=1}^n  \frac 1{{\chi_i}} \right)\ge n^2.
$$
\end{fact}
\bp
This is another way to express  the fact that the arithmetic mean is greater than or equal to   the harmonic mean.
\ep

In view of Proposition \ref{prop:fps}, there is at most one solution to the fixed-point equation $v=f(u,v)$ over $[v_0(u),+\infty)$, and this solution $v(u)$ exists  if and only if 
\eq{\label{eq:cuni}
 f(u,v_0(u))\le v_0(u).
}
Moreover, the solution can  be found by a simple binary search algorithm.

Now let $g$ be the function defined 
by
	\begin{align*}
	\forall u, \forall v\ge v_0(u), \quad g_i(u,v) = \frac{1}{2} + \frac{n}{4} \left(\sqrt{ \delta_i(u,v)}  - \sqrt{v}\right).
	\end{align*}
For any $u$ that satisfies \eqref{eq:cuni}, we define $\phi(u)=g(u,v(u)).$
	
\begin{proposition}\label{prop:fp}
The unique solution to the fixed-point equation $v = f(a,v)$ is $v(a)=(1-2q)^2$. Moreover,  we have $v(a)>v_0(a)$ and $p=\phi(a)$.
\end{proposition}
\bp
Let $v=(1-2q)^2$. It can be readily verified from  \eqref{eq:trinome} that $p_i=g_i(a,v)$. It then follows from Assumption \ref{a:fixedpoint} that $\delta_i(a,v)>0$ and thus $v> v_0(a)$.
Moreover,
\begin{align*}
v =  \left(1 - \frac{2}{n} \sum_{i=1}^n p_i\right)^2
=  \left(1 - \frac{2}{n} \sum_{i=1}^n g_i(a,v)\right)^2
=\left(\frac 1 2 \sum_{i=1}^n  \sqrt{\delta_i(a,v)} -\frac n 2 \sqrt{v}\right)^2,  
\end{align*}
so that, taking the square root of both terms, $v$ satisfies the fixed-point equation $v=f(a,v)$. This shows   that $v(a)=v$ and $p=g(a,v(a))=\phi(a)$.
\ep

\subsubsection*{Estimator}

Proposition \ref{prop:fp} suggests that it is sufficient to estimate $a$ in order to retrieve $p$.
 We propose  the following estimate of $a$,
\eq{\label{eq:Ncalculation}
{\hat a}_i(t) = \frac{t-1}{t} {\hat a}_i(t-1)  + \frac{1}{t(n-1)\alpha^{2}}  \sum_{j \neq i} { \indic\{X_i(t) X_j(t) = 1 \}},
}
with ${\hat a}_i(0) = 0$ for all $i=1,\ldots,n$. Note  that
\eq{\label{eq:adef}
{\hat a}_i(t) = \frac{1}{t(n-1)\alpha^2} \sum_{s= 1}^t \sum_{j \neq i} { \indic\{X_i(s) X_j(s) = 1 \}},
}
so that ${\hat a}_i(t)$ is the empirical average of the number of labellers whom agree with $i$ for tasks $1,\ldots,t$.  We use the definition \eqref{eq:Ncalculation} to highlight the fact that ${\hat a}(t)$ can be computed in a streaming fashion. 

The time complexity of the update \eqref{eq:Ncalculation} is $O(n^2)$ per task. Using the fact that $ \indic\{x= 1 \}=\frac 1 2 (x+|x|)$ over $\{-1,0,1\}$, we can in fact update the estimator $\hat a(t)$ as follows,
$$
{\hat a}_i(t) = \frac{t-1}{t} {\hat a}_i(t-1)  + \frac{X_i(t)S(t)+|X_i(t)| (|N(t)|-2)}{2t(n-1)\alpha^{2}},
$$
where  $S(t)=\sum_{j=1}^n X_j(t)$ is the sum of the labels of task $t$ and $N(t)=\sum_{j=1}^n |X_j(t)|$ is the total number of actual labellers for task $t$. The time complexity of the update is then $O(n)$ per task.

\subsubsection*{Algorithm}

Given this estimation of the vector   $a$ of agreement rates, 
our  estimation of the vector ${p}$ of error probabilities is 
\begin{itemize}
\item ${\hat p}(t) = \phi(\hat a(t))$ if the fixed-point equation $v = f({\hat a}(t),v)$ has a unique solution,
\item  ${\hat p}(t) = (\frac{1}{2},...,\frac{1}{2})$ otherwise.
\end{itemize}
We  denote by $\hat w(t)$ the corresponding weight vector, with  $\hat w_i(t)=\log (1/\hat p_i(t)-1)$ for all $i=1,\ldots,n$. These weights inferred from tasks $1,\ldots,t$ are  used to label task $t+1$ according to weighted majority vote, as defined by \eqref{eq:wmv}. We refer to this algorithm as the agreement-based (AB) algorithm.

\section{Performance guarantees}

\label{sec:perf}

In this section, we provide  performance guarantees for the AB  algorithm, both in terms of statistical error and computational complexity, and show that its cumulative regret compared to an oracle that knows the latent parameter $p$ is finite, for any number of tasks.

\subsection{Accuracy of the estimation}

Let $\gamma =v(a) - v_0(a)$.  This is a fixed parameter of the model. Observe that $\gamma \in (0,1]$ in view of Proposition \ref{prop:fp} and the fact that $v(a)=(1-2q)^2\le 1$. Theorem~\ref{th:concentration}, proved in 
the Appendix, gives a  concentration inequality on the  estimation error 
at time $t$ (that is, after having processed tasks $1,\ldots,t$). 
We denote by $||\cdot||_\infty$  the $\ell_\infty$ norm in $\RR^n$.
	
\begin{theorem}\label{th:concentration}
For any  $ \varepsilon \in (0,\frac 1{20}]$,
\begin{align*}
\PP(||\hat p(t)-p||_\infty\geq \varepsilon)&\leq  2n\exp\left( - \frac{  \gamma^3   \alpha^4}{ 8 }  t \varepsilon^2\right).
\end{align*}
\end{theorem}

\begin{corollary}\label{cor:concentration}
The estimation error is of order
\begin{align*}
||\hat p(t)-p||_\infty &= O\left(\frac 1  { \gamma^{3 \over 2}\alpha^2 } \sqrt{ \log n  \over t} \right).
\end{align*}
\end{corollary}
As shown by Corollary~\ref{cor:concentration}, Theorem~\ref{th:concentration} yields the error rate of our algorithm in the regime where $q$ and $\alpha$ are fixed and $t/\log n \to \infty$, but is much stronger than what one may obtain through an asymptotic analysis. Indeed, for any values of $n$ and $t$, Theorem~\ref{th:concentration} shows that the mean estimation error  exhibits sub-Gaussian concentration, and directly yields confidence regions for the vector ${\hat p}(t)$. This may useful for instance in a slightly different setting where the number of samples $n t$ is not fixed, and one must find a stopping criterion ensuring that the estimation error is below some target accuracy. An example of this setting arises when one attempts to identify the best $k< n$ labellers under some constraint on the number of samples.

\subsection{Complexity}
In order to calculate ${\hat p}(t)$, we only need to store the value of ${\hat a}(t)$, which requires $O(n)$ memory space. Further, we have seen that the update of $\hat a(t)$ requires $O(n)$ operations. For any $\epsilon(t)> 0$ computing the fixed point $v(\hat a(t))$ (using a binary search) up to accuracy $\epsilon(t)$ requires $O(n \log(1/\epsilon(t)))$ operations. The accuracy of our estimate  is  $O(\sqrt{\log n /t })$ (omitting the factors $\alpha$ and $\gamma$), so that one should use $\epsilon(t) = O(\sqrt{\log n /t })$. The time complexity of our algorithm is then $O(n \log t)$. It is noted that any estimator of $p$ requires at least $O(n)$ space and $O(n)$ time, since one has to store at least one statistic per labeller, and each component of $p$ must be estimated. Therefore the complexity of the AB  algorithm is optimal (up to  logarithmic factors) in  both  time and space. 

\subsection{Regret}

The regret is a  performance metric that allows one to compare  any algorithm to   the optimal decision knowing    the latent parameter $p$, given by some oracle.
We define two notions of regret.  The {\it simple} regret is the difference between the prediction error rate of our algorithm and that of the optimal decision for  task $t$. By Proposition~\ref{pr:weightedmajority}, the optimal decision follows from    weighted majority voting with weights $w$ given by the oracle;   we denote by ${G^\star}(t)$ the corresponding output for task $t$. The  simple regret is then
\eqs{
r(t) = \PP( {\hat G}(t) \neq G(t))- \PP({G^\star}(t) \neq G(t)).
}
The second performance criterion is the {\it cumulative} regret, $
R(t) = \sum_{s=1}^t r(s)$, that is the difference between the expected number of errors done by our algorithm and that of the optimal decision, for tasks $1,\ldots,t$. 

Let $\eta = \min_i p_i(1-p_i)$ and
$
\lambda =  \min_{x\in \{-1,0,1\}^n: w^Tx\ne 0} |w^Tx |.
$ The following result, proved in the Appendix, shows that the cumulative regret of the AB algorithm is finite.

\begin{theorem}\label{th:regret}
Assume that $\eta > 0$. We have
\begin{align*}
	r(t) &\leq  2n \exp\left({-\frac{\gamma^3\alpha^4c^2}{8}t }\right),
\end{align*}
with $c=\frac 1 4 \min(\lambda\eta,\frac 1 5)$, and
\eqs{
	R(t) \leq  \frac {16 n}{\gamma^3\alpha^4c^2}.
}
\end{theorem}

\section{Non-Stationary Environment}
\label{sec:nonstat}

We have so far assumed a stationary environment so that the latent parameters $p$ stay constant over time. We shall see  that, due to its  streaming nature, our algorithm  is also well-suited to non-stationary environments. In practice, the vector of error probabilities $p$ may vary over time  due to several reasons, including:
\begin{itemize}
\item{\bf Classification needs:} The type of data to  label may change over time depending on the customers  of crowdsourcing and the market trends. 
\item{\bf Learning:} Most tasks (e.g., recognition of patterns in images, moderation tasks, spam detection) have a learning curve, and labellers become more reliable as they label more tasks. 
\item{\bf Aging:} Some tasks require knowledge about the current situation (e.g., recognizing trends, analysis of the stock market) so that highly reliable labellers may become less accurate if they do not keep themselves up to date.
\item{\bf Dynamic population:} The population of labellers may change over time. While we assume that the total number of labellers is fixed, some labellers may periodically leave the system and be replaced by new labellers.
\end{itemize}

\subsection{Model and algorithm}

We assume that  the number of labellers $n$ does not change over time but  that $p$ varies with time at speed $\sigma$, so that for each labeller $i\in \{1,\ldots,n\}$, 
\eqs{
	| p_i(t) - p_i(s) | \leq \sigma |t - s| \;\;\;,\;\;\; \forall t,s\ge 1.
}

We propose to adapt our algorithm  to  non-stationary environments by replacing empirical averages with exponentially weighted averages. Specifically, given $\beta\in (0,1)$ an averaging parameter, we define the estimate ${\hat a}^{\beta}(t)$ of the vector $a(t)$ of agreement rates at time $t$ by
\eq{\label{eq:Nnonstat}
{\hat a}^{\beta}_i(t) = (1 - \beta){\hat a}^{\beta}_i(t-1)  + \beta \frac{X_i(t)S(t)+|X_i(t)| (|N(t)|-2)}{2(n-1)\alpha^{2}}.
}
with ${\hat a}^{\beta}_i(0) = 0$ for all $i=1,\ldots,n$. As in the stationary case, the estimate   ${\hat a}^{\beta}(t)$ can be calculated as a function of  ${\hat a}^{\beta}(t-1)$ and the sample $X(t)$ in $O(n)$ time, which fits the streaming setting. One may readily check that:
\eq{\label{eq:abel}
{\hat a}^{\beta}_i(t) =  \sum_{s = 1}^t { \beta(1-\beta)^{t -s} \over (n-1)\alpha^2} \sum_{j \neq i} { \indic\{X_i(s) X_j(s) = 1 \}}.
}

\subsection{Performance guarantees}
As in the stationary case, we derive concentration inequalities. Observe that the parameter $\gamma$ now varies over time. The proof of Theorem \ref{th:concentrationnonstat} is given in the appendix.
\begin{theorem}\label{th:concentrationnonstat}
Assume that ${2 \sigma \over \beta} \leq {\gamma(t) \over 80}$. Then for all $ \epsilon \in (0,{\gamma(t) \over 80} - {2 \sigma \over \beta}]$, 
\als{
\PP\Big( ||{\hat p}(t)-p(t)||_\infty \geq \frac 4 {\gamma(t)^{{3 \over 2}}}(\epsilon + 2 \frac \sigma \beta) \Big) 
   \leq 2 n \exp\left( - \frac {2 \epsilon^2 \alpha^4 }\beta \right).
}
\end{theorem}
\begin{corollary}\label{cor:concentrationnonstat}
  The estimation error  is of order :
  $$||{\hat p}(t)-p(t)||_\infty= O\left( \frac 1 { \gamma(t)^{{3 \over 2}}} \left(   \frac{\sqrt{\beta \log n}}{\alpha^2}  + { \sigma \over \beta} \right) \right).$$ 
\end{corollary}

The expression of the estimation error  shows that choosing $\beta$ involves a bias-variance tradeoff, where the variance term is proportional to $\sqrt{\beta}$ and the  bias term is proportional to $1/\beta$. We derive the order of the optimal value of $\beta$ minimizing the estimation error of our algorithm. This is of particular interest in the slow-variation regime $\sigma \to 0^+$, since in most practical situations the environment  evolves slowly (e.g., at the timescale of hundreds of tasks). 
\begin{corollary}\label{cor:window}
Letting  $\beta = \alpha^{4 \over 3} \sigma^{2 \over 3} /(\log n)^{3}$,   the estimation error is
of order $$||{\hat p}(t)-p(t)||_\infty= O\left( \frac{\sigma^{1 \over 3} (\log n)^3}{ \alpha^{ {4 \over 3}} \gamma(t)^{{3 \over 2}}} \right).$$
\end{corollary}

\section{Numerical Experiments}
\label{sec:num}
In this section, we investigate the performance of our Agreement-Based (AB) algorithm on both  synthetic data, in  stationary and non-stationary environments, and real-world datasets.

\subsection{Stationary environment}
We start with synthetic data in  a stationary environment. We consider a generalized version of the hammer-spammer model with an even number of labellers $n$,   the first half of the labellers being identical and informative and the second half of the labelers being non-informative, so that $p_i = p_1 < {1 \over 2}$ for $i\in\{1,\ldots, {n \over 2}\}$ and $p_{i} = {1 \over 2}$ otherwise.

\begin{figure}[ht]
\begin{center}
\subfigure[Maximum estimation error $||\hat p(t)-p||_\infty$.]{\includegraphics[width=\figsize]{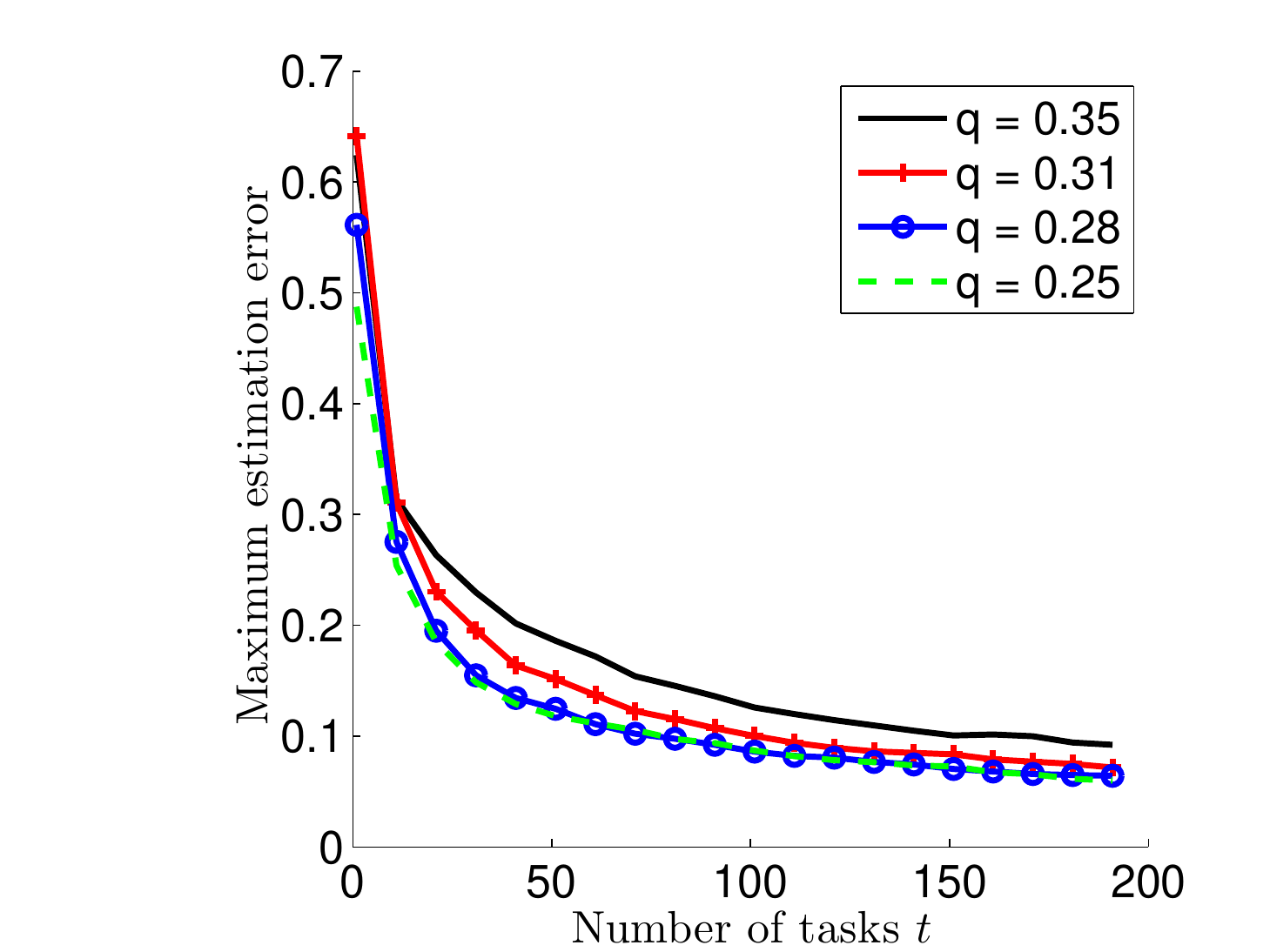}}
\subfigure[Average estimation error $\frac 1 n||\hat p(t)-p||_1$]{\includegraphics[width=\figsize]{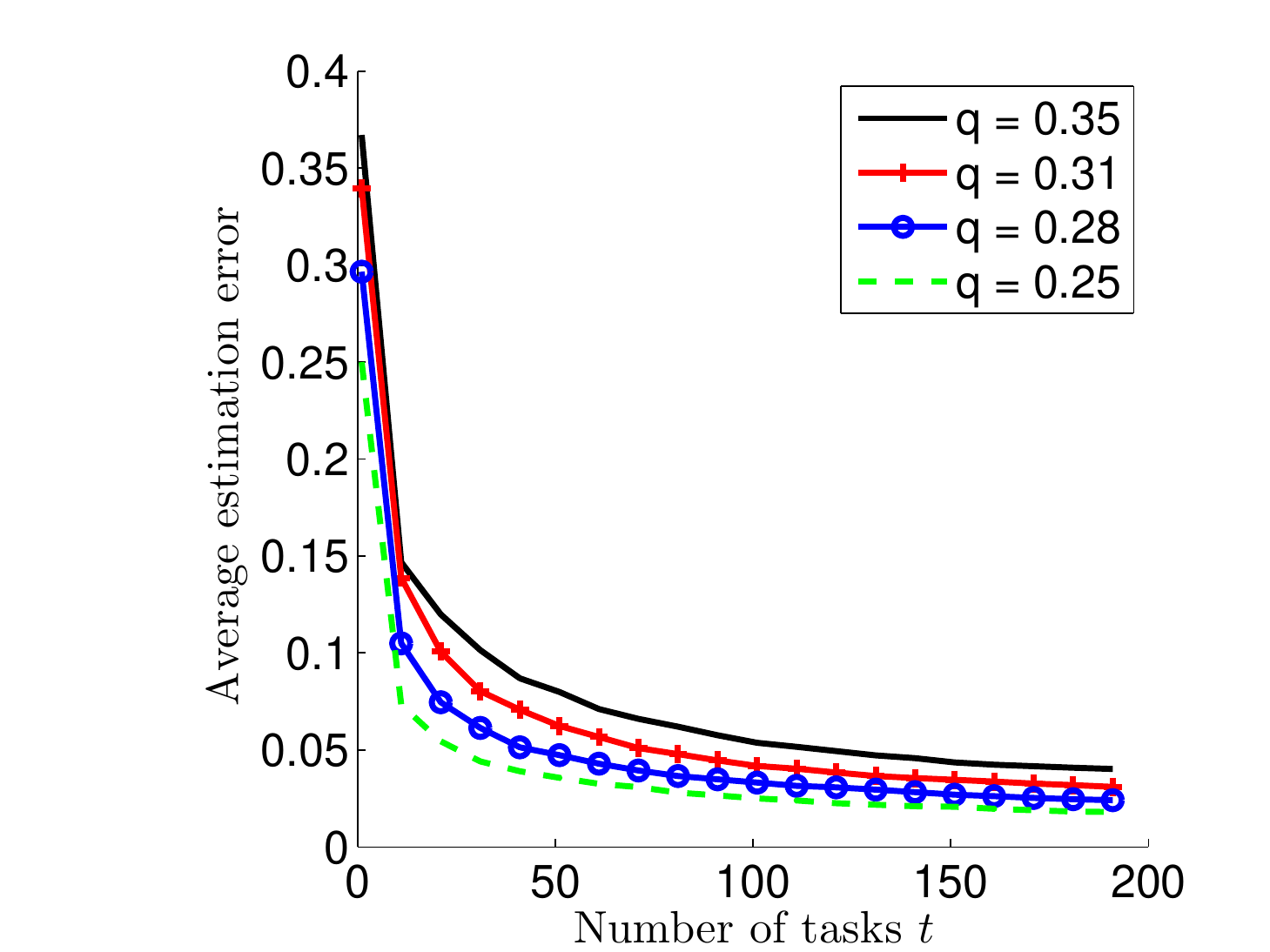}}
\caption{Estimation error  with respect to the number of tasks $t$.}
\label{fig:error_t}
\end{center}
\end{figure}

Figure~\ref{fig:error_t}  shows  the  estimation error  on  $p$ with respect to the number of tasks $t$. There are  $n=10$ labellers, all working on all tasks (that is $\alpha = 1$) and various values of the average error probability $q$. The error is decreasing with $t$ in $O(1/\sqrt{t}$) and increasing with $q$, as expected: the problem becomes harder as $q$ approaches $1 \over 2$, since labellers become both less informative and less correlated.

\begin{figure}[ht]
\begin{center}
\includegraphics[width=\figsize]{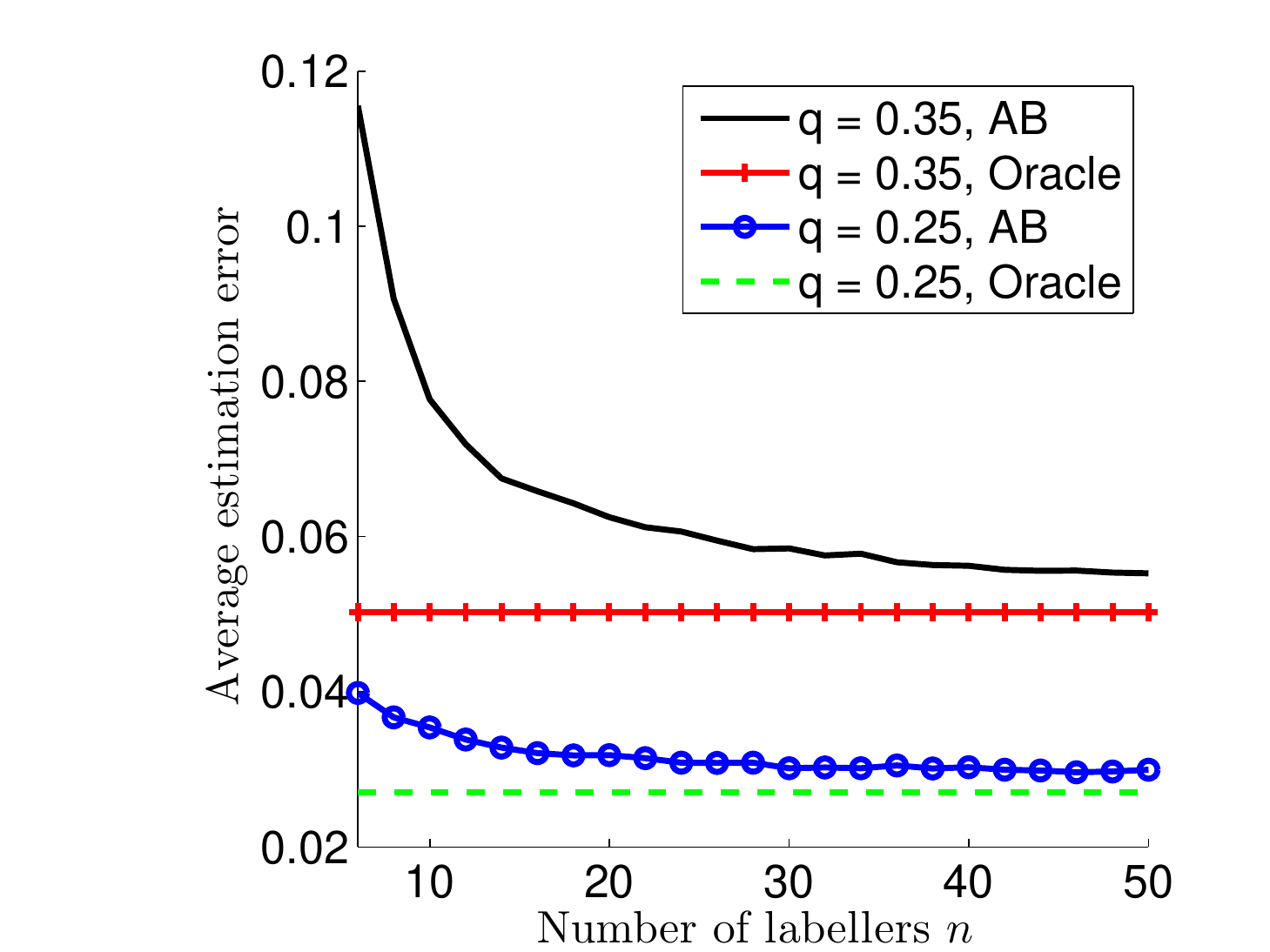}
\end{center}
\caption{Average estimation error $\frac 1 n||\hat p(t)-p||_1$ with respect to the number of labellers $n$.}
\label{fig:average_error_n}
\end{figure}

Figure~\ref{fig:average_error_n} shows the average estimation error of our algorithm for $t=50$ tasks as a function of the number of labellers $n$. We compare our algorithm  with an oracle which knows the values of the truth $G(1),\dots,G(t)$ (note that this is  different from the oracle used to define the regret, which knows the parameter $p$ and must guess the truth  $G(1),\dots,G(t)$). This estimator (which is optimal) simply estimates   $p_i$ by the empirical probability that labeller $i$ disagrees with the truth. Interestingly, when $n$ increases, the error of our algorithm approaches that of the oracle, showing that our algorithm  is nearly optimal. 

\begin{figure}[ht]
\begin{center}
\subfigure[Maximum estimation error $||\hat p(t)-p||_\infty$]{\includegraphics[width=\figsize]{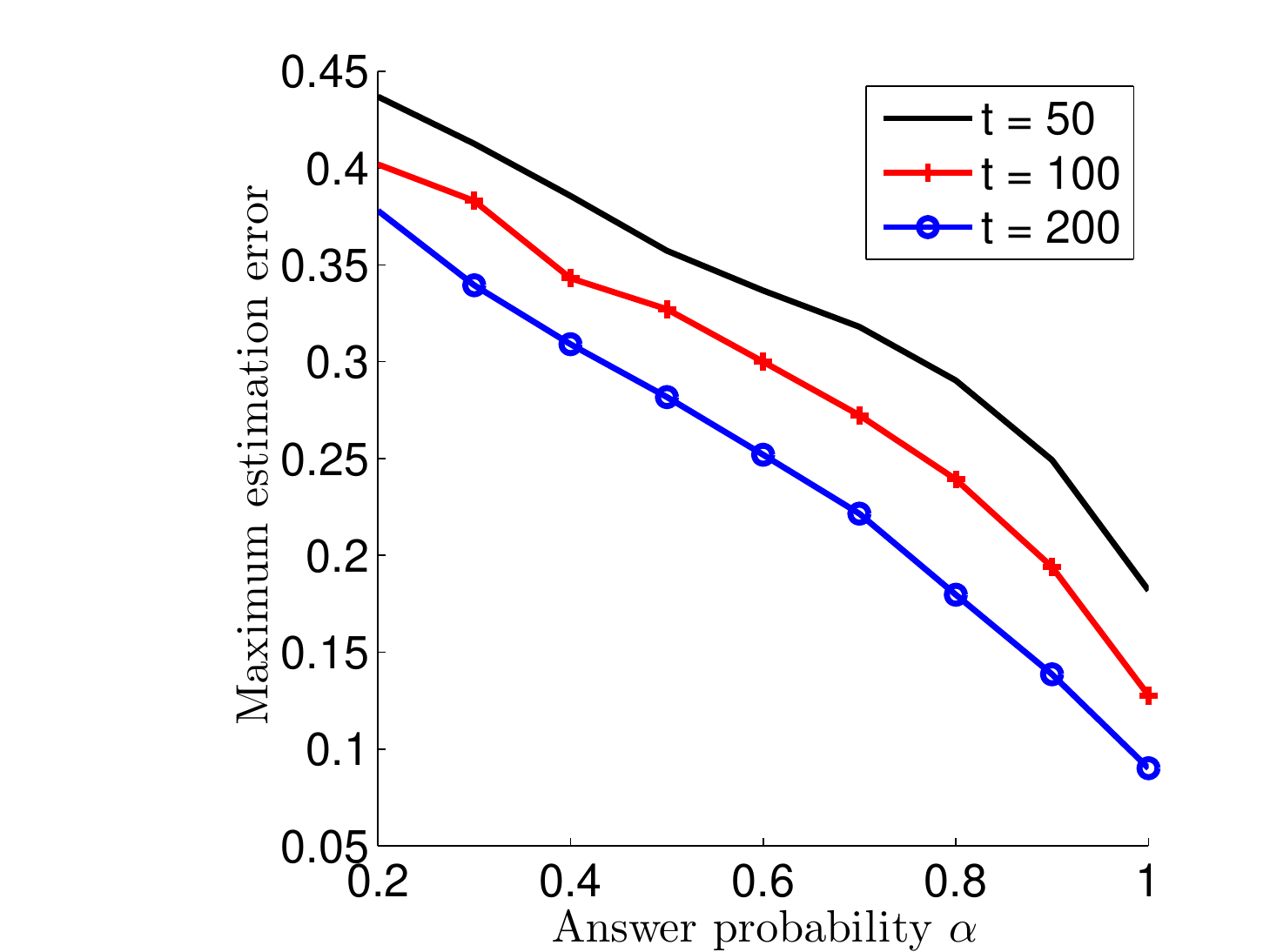}}
\subfigure[Average estimation error $\frac 1 n||\hat p(t)-p||_1$]{\includegraphics[width=\figsize]{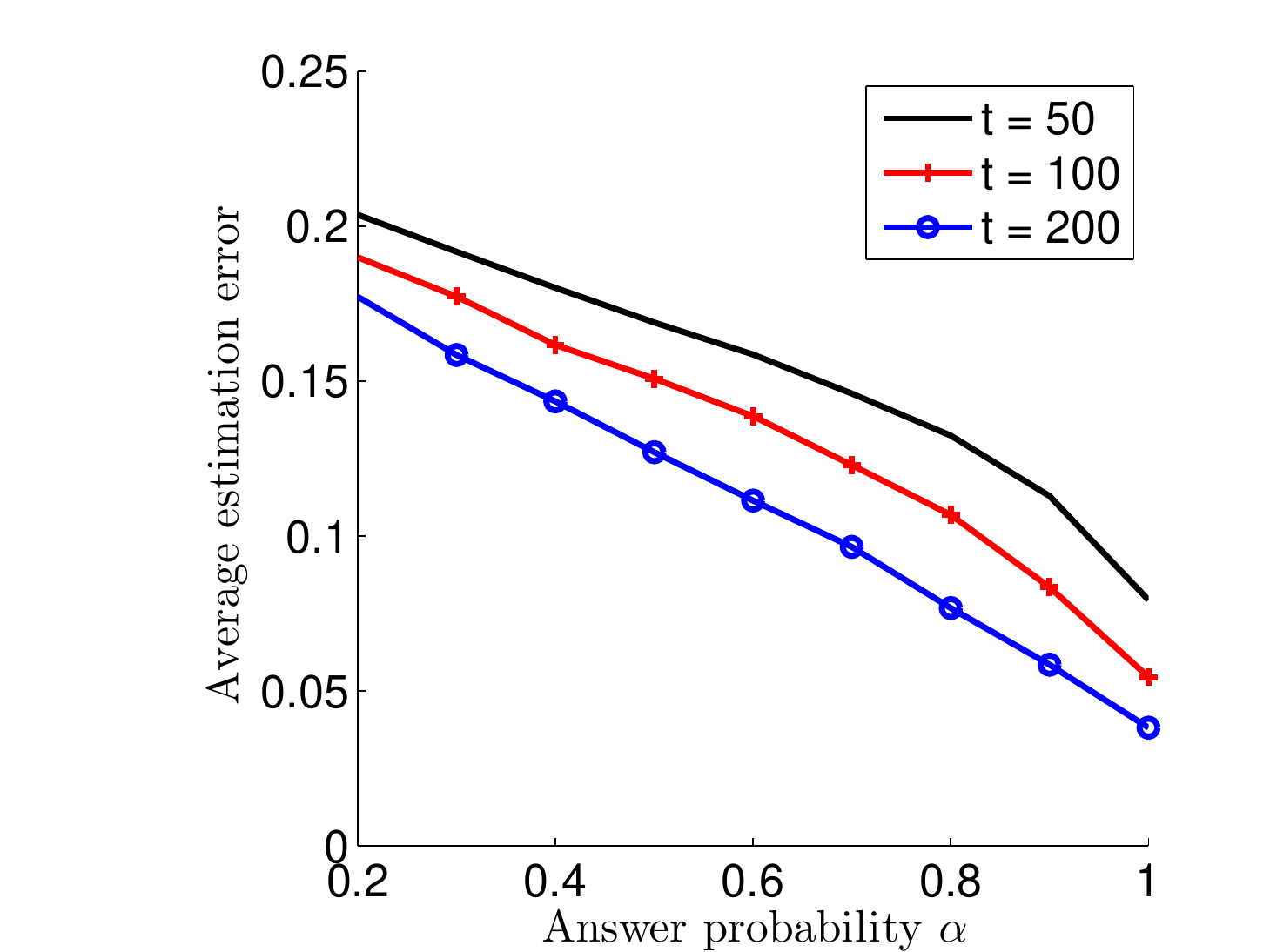}}
\caption{Estimation error with respect to the anwser probability $\alpha$.}
\label{fig:error_alpha}
\end{center}
\end{figure}

On Figure \ref{fig:error_alpha} we present the impact of the answer probability $\alpha$ on the estimation error, for $n=10$ labellers. As expected, the estimation error decreases with $\alpha$. The dependency is approximately linear, which suggests that our upper bound on the estimation error given in Corollary \ref{cor:concentration}, which is inversely proportional to $\alpha^2$, can be improved.

\begin{figure}[ht]
\begin{center}
\includegraphics[width=\figsize]{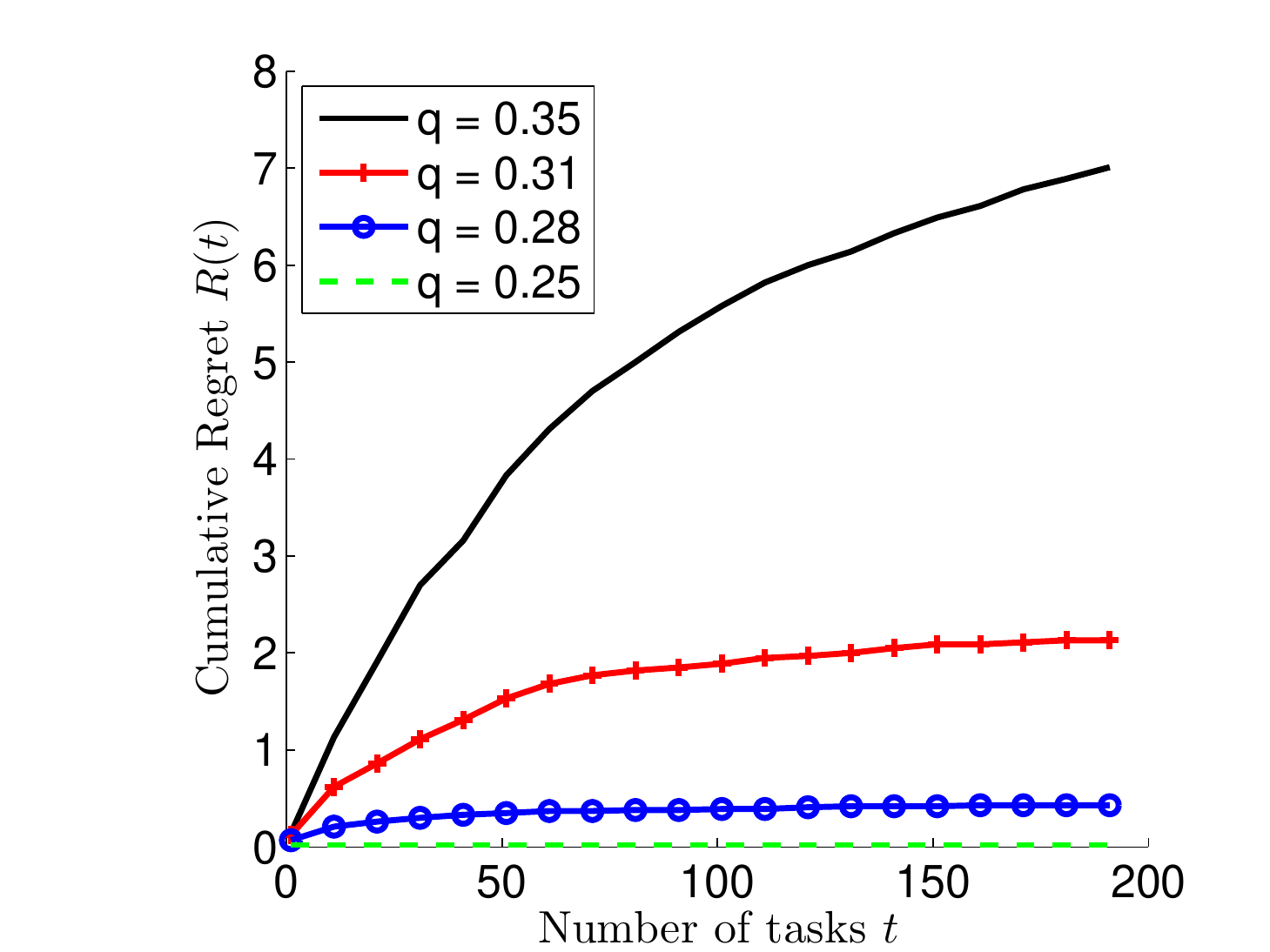}
\end{center}
\caption{Cumulative regret $R(t)$ with respect to the number of tasks $t$.}
\label{fig:regret_t}
\end{figure}

On Figure~\ref{fig:regret_t} we present the cumulative regret  $R(t)$ with respect to the number of tasks $t$, for $n=10$ labellers and different values of  the average error probability $q$. As for the estimation error, the cumulative regret increases with $q$, so that the problem becomes harder as $q$ approaches $1 \over 2$, as expected. We know from Theorem \ref{th:regret} that this cumulative regret is finite, for any $q$ that satisfies Assumption 1 (here, $q<0.4$). 
We observe that this regret is suprisingly low: for $q=0.25$, the cumulative regret is close to 0, meaning that there is practically no difference with the oracle, which knows perfectly the parameter $p$; for $q = 0.31$, our algorithm  makes less than   $2$ prediction errors on average compared to the oracle.
 
\subsection{Non-stationary environment}

We now turn to non-stationary environments. We assume that the error probability of each labeller evolves as a sinusoid between $0$ and ${1 \over 2}$ with some common  frequency $\omega$, namely $p_i(t) = {1 \over 4}(1 + \sin (\omega t + \varphi_i))$. The phases are  regularly spaced over $[0,2\pi]$, i.e., $\varphi_i=2\pi (i/n)$ for all $i=1,\ldots,n$. 

\begin{figure}[ht]
\begin{center}
\includegraphics[width=\figsize]{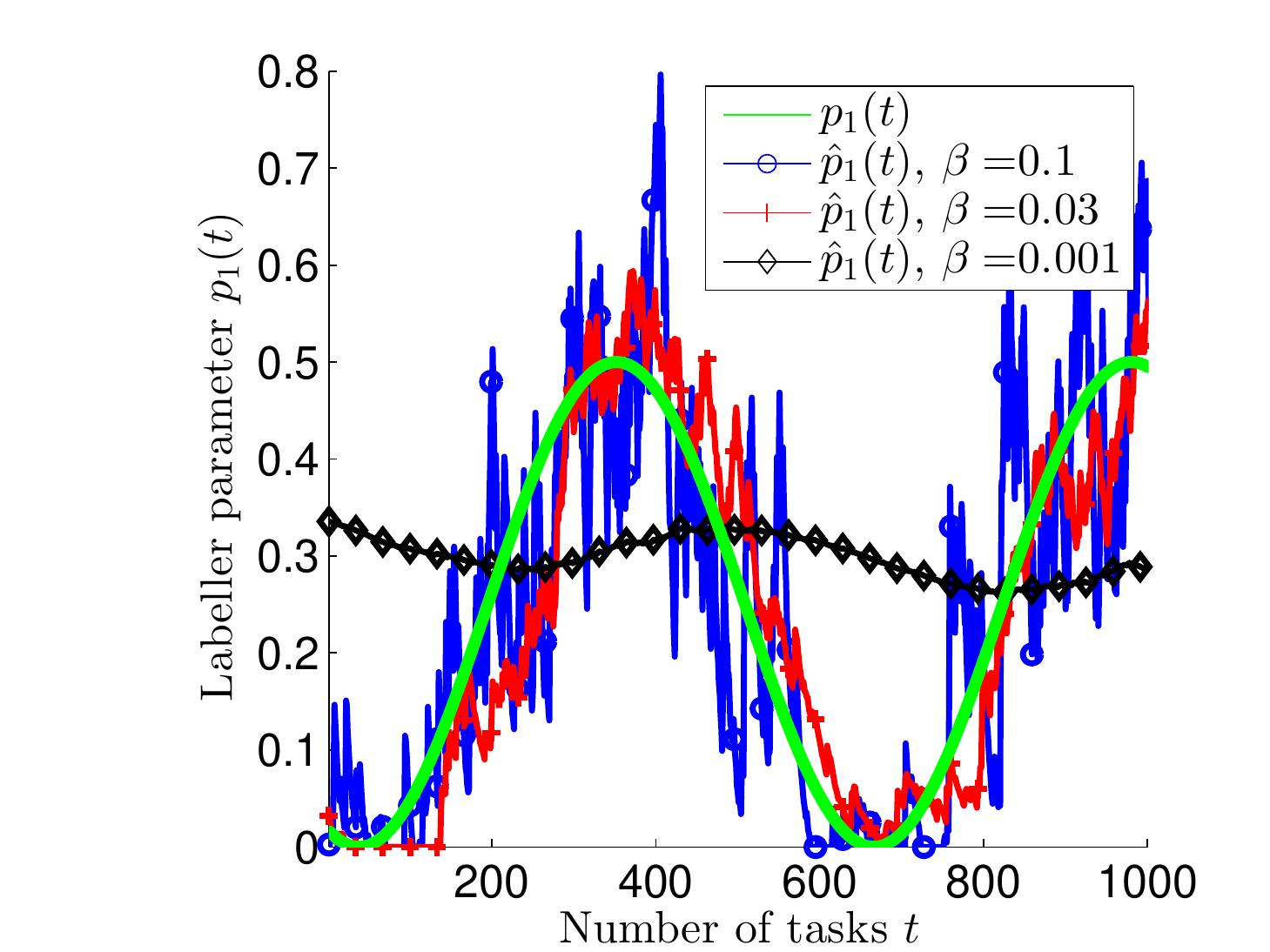}
\end{center}
\caption{Estimate of $p_1(t)$ with respect to the  number of tasks $t$, non-stationary environment.}
\label{fig:nonstationary}
\end{figure}

Figure~\ref{fig:nonstationary} shows the true parameter $p_1(t)$ of labeller 1 and the estimated value $\hat p_1(t)$ on a sample path for $n = 10$ labellers, $\omega = 10^{-2}$ and various values of the averaging parameter $\beta$. One clearly sees the bias-variance trade-off underlying the choice of $\beta$: choosing a small $\beta$ yields small fluctuations but poor tracking performance, while $\beta$ close to $1$ leads to large fluctuations centered around the correct value. Furthermore, the natural intuition that $p_1(t)$ is harder to estimate when it is close to $1 \over 2$ is apparent. Finally, for $\beta$ properly chosen (here $\beta=0.03$), our algorithm effectively tracks the evolving latent parameter $p_1(t)$. 

\begin{figure}[ht]
\begin{center}
\includegraphics[width=\figsize]{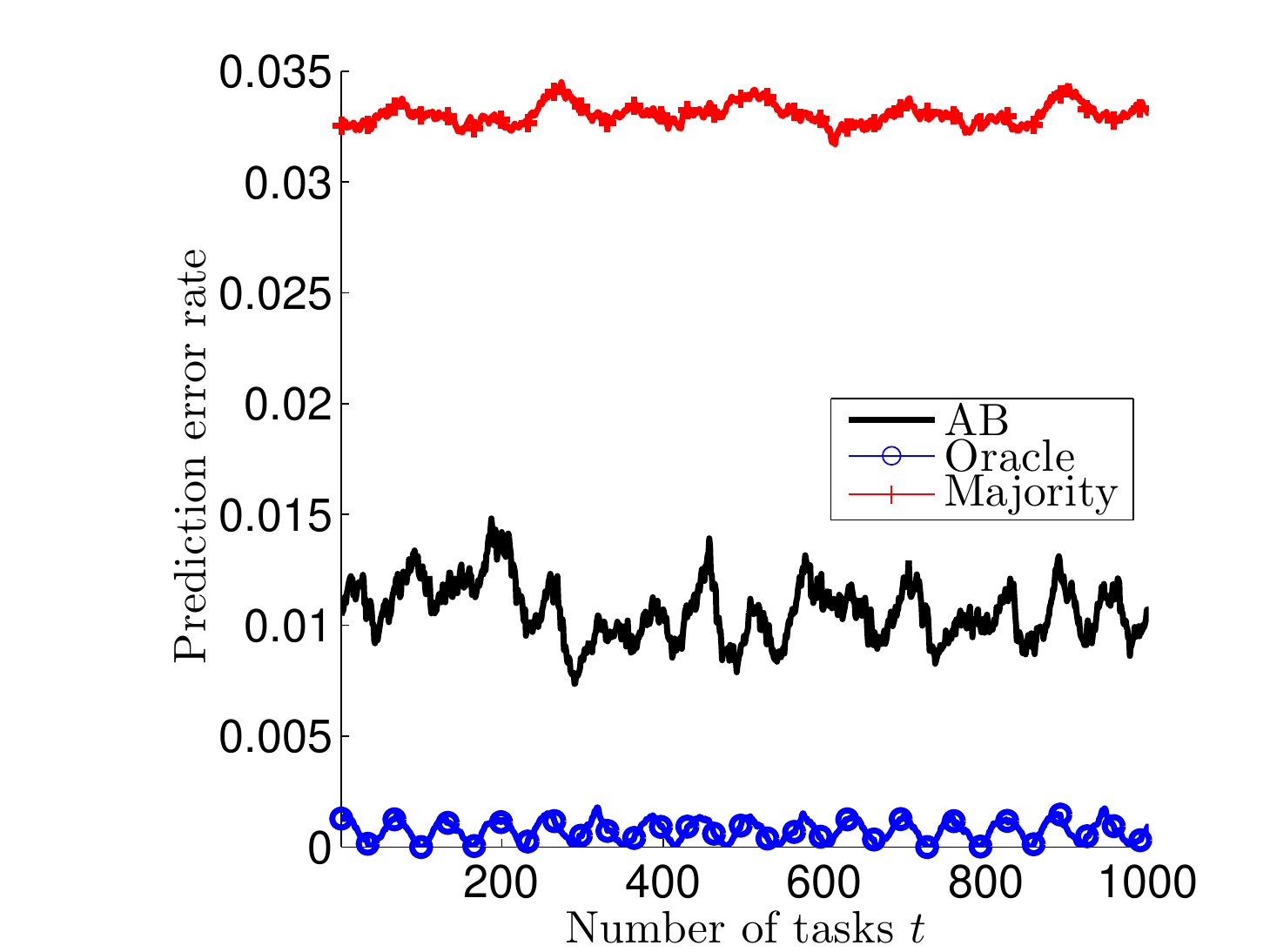}
\end{center}
\caption{Prediction error rate with respect to the  number of tasks $t$, non-stationary environment.}
\label{fig:nonstationary2}
\end{figure}

Figure \ref{fig:nonstationary2} shows the  prediction error rate of our algorithm, for $\beta=0.03$, compared to that of majority vote and to that of an oracle that known $p(t)$ exactly for all tasks $t$. 

\subsection{Real datasets}

Finally, we test the performance of our algorithm on real, publicly available datasets (see \cite{WWB09,ZL15} and references therein), whose main characteristics are summarized in Table \ref{tab:dataset}. 
When the  data set has more than two possible labels (which is the case of the ``Dog" and the ``Web" datasets), say in the set $\{1,\ldots,L\}$, we merge all labels $\ell\le L/2$ into label $+1$  and all labels $\ell >L/2$ into label $-1$. 

\begin{table}[ht]
\begin{center}
\begin{tabular}{lccc} \hline
{\bf Dataset} &   \# Tasks & \# Workers  & \# Labels \\ \hline 
{\bf Bird} & 108 & 39 & 4,212 \\ 
{\bf Dog} & 807 & 109 & 8,070 \\
{\bf Duchenne} & 160 & 64 & 1,311 \\
{\bf Rte} & 800 & 164 & 8,000 \\
{\bf Temp} & 462  & 76  & 4,620 \\ 
{\bf Web} & 2,665 & 177 & 15,567 \\ \hline
\end{tabular}
\end{center}
\caption{Summary of the considered datasets.}
\label{tab:dataset}
\end{table}

Each dataset contains the ground-truth of each task, which allows one to assess the prediction error rate of any algorithm. The results are reported in Table~\ref{fig:real_data} for the following algorithms:
\begin{itemize}
\item Majority Vote (MV),
\item a standard Expectation Maximization (EM) algorithm known as the DS estimator \cite{DS79},
\item our Agreement-Based (AB) algorithm.
\end{itemize}
Except for the ``Temp'' dataset, our algorithm  yields some improvement compared to MV, like EM,  and a significant  performance gain  for the ``Web'' data set, for which more samples are available. 
The performance of AB and EM are  similar for all datasets except for  ``Bird", where the number of tasks is limited; this is remarkable given the much lower    computational cost of AB, which is linear in the number of samples. 

\begin{table}[ht]
\begin{center}
\begin{tabular}{lccc} \hline
{\bf Dataset} &  MV & EM & AB \\ \hline 
{\bf Bird} & 0.24 & 0.10 & 0.23\\ 
{\bf Dog} & 0.00 & 0.00 & 0.00 \\ 
{\bf Duchenne} & 0.28 & 0.28 &  0.26 \\ 
{\bf Rte} & 0.10 & 0.07 & 0.08 \\
{\bf Temp} & 0.06 & 0.06 & 0.07 \\ 
{\bf Web} & 0.14 & 0.06  & 0.06 \\  \hline
\end{tabular}
\end{center}
\caption{Prediction error rates of different algorithms on real datasets.}
\label{fig:real_data}
\end{table}

\section{Conclusion}
\label{sec:conc}

We have proposed a  streaming algorithm for performing crowdsourced data classification. The main feature of this algorithm is to adopt a ``direct approach'' by inverting the relationship between the agreement rates $a$ between various labellers and the latent parameter $p$. This Agreement-Based (AB) algorithm is not a spectral algorithm and does not require to store the task-labeller matrix. Apart from a simple line search, AB does not involve an iterative scheme such as EM or BP. 

We have provided performance guarantees for our algorithm in terms of  estimation errors. Using this key result, we have shown that our algorithm is optimal  in terms of  both time complexity (up to logarithmic factors) and regret (compared to the optimal decision). Specifically, we have proved that  the   cumulative regret is finite, independently of the number of tasks; as a comparison, the cumulative regret of a basic algorithm based   on majority vote increases {\it linearly} with the number of tasks. We have assessed the performance of AB on both synthetic and real-world data; for the latter, we have seen that AB generally behaves like EM, for a much lower time complexity.

 We foresee two directions for future work: on the theoretical side, we want to investigate the extension of AB to more intricate models featuring non-binary labels and where the error probability of labellers depends on the considered task. We would also like to extend our analysis to the sparse regime  considered in \cite{KOS11}, where the number of answers on a given task does not grow with $n$, so that $\alpha$ is proportional to $1/n$. On the practical side, since AB is designed to work with large data sets provided in real-time as a stream, we hope to be able to experiment its performance on a real-world system.

\hspace{1cm}

\bibliographystyle{abbrv}
\bibliography{crowdsourcing_streaming}  

\newpage
\appendix

\section{Proof of Theorem \lowercase{\ref{th:concentration}}}

We denote by $||\cdot||_1$ and $||\cdot||_\infty$ the $\ell_1$ norm and the $\ell_\infty$ norm in $\RR^n$, respectively.

\subsection{Outline}
The proof consists of  three steps:
\begin{enumerate}
\item {\bf Concentration of ${\hat a}(t)$.} Using 
Hoeffding's inequality, we prove a concentration inequality on $\hat a(t)$.
\item {\bf Fixed-point uniqueness.} From the concentration of ${\hat a}(t)$, we deduce that $v({\hat a}(t))$ concentrates around $v(a)$, so that  the fixed-point equation $v=f(\hat a(t),v)$ has a unique solution with high probability.
\item {\bf Smooth dependency between ${\hat a}(t)$ and ${\hat p}(t)$.} When a unique fixed point exists, the mapping ${\hat a}(t) \mapsto {\hat p}(t)$ depends smoothly on each component of ${\hat a}(t)$, which implies the concentration of $\hat p(t)$.
\end{enumerate}

\subsection{Intermediate results}

Recall that \eqref{eq:cuni} is  a necessary and sufficient condition for the existence and uniqueness of a solution to the fixed-point equation $v=f(u,v)$.
Proposition \ref{pr:existancecondition} provides a simpler, sufficient condition.
For any  $u \in [0,1]^n$, let
\eqs{
v_1(u) =\frac 2 {n}  \sum_{i=1}^n (2 {u}_i - 1).
} 

\begin{proposition}\label{pr:existancecondition}
If $ v_1(u)\ge  v_0(u)$ then there is  a unique solution to the fixed-point equation $v=f(u,v)$. 
\end{proposition}
\bp 
By the  Cauchy-Schwartz inequality,
 $$\sum_{i=1}^n  \sqrt{\delta_i(u,v)} \leq \sqrt{n \sum_{i=1}^n \delta_i(u,v) },$$ 
 so that for all $v>v_0(u)$,
\als{
	f(u,v) &\leq \frac{n}{(n-2)^2} \sum_{i=1}^n \delta_i(u,v), \sk
		&=\frac 1{(n-2)^2}\left( {n^2} v - \frac{4(n-1)}{n} \sum_{i=1}^n (2u_i -1)\right), \sk
		&= \frac{n^2 v - 2 v_1(u)(n-1)}{(n-2)^2}.
}
In particular,
\eqs{
f(u,v) - v \leq 2\frac{n-1}{(n-2)^2} ( v - v_1(u)). 
}
If $v_1(u) \ge  v_0(u)$, then $f(u,v_0(u))\le v_0(u)$ and there is a  unique solution to the fixed-point equation $v=f(u,v)$.
\ep

Proposition \ref{pr:lowerboundalpha}  will be used to  prove that the fixed-point equation $v=f(u,v)$ has a unique solution for any $u$ in some neighborhood of $a$.

\begin{proposition}\label{pr:lowerboundalpha}
We have $v_1(a) - v_0(a) > v(a)$.
\end{proposition}
\bp
By the definition of $a$,
\als{
	(n-1)\sum_{i=1}^n a_i &= \sum_{i\ne j} (p_i p_j + (1-p_i)(1-p_j)) \sk
		&=  \left(\sum_{i=1}^n p_i\right)^2 +  \left(n - \sum_{i=1}^n p_i \right)^2
		- \sum_{i=1}^n (p_i^2 + (1-p_i)^2).
}
Using the fact that $p_i^2 + (1-p_i)^2 \leq \frac 12$ for all $p_i \in [0,1]$ and  $\sum_{i=1}^n p_i = nq$, we obtain the lower bound:
\eqs{
(n-1)\sum_{i=1}^n a_i \geq  \frac{n}{2}( n (1-2q)^2 + n - 1).
}
In particular, 
\eqs{
v_1(a) = \frac 2 {n}\sum_{i=1}^n (2 a_i - 1) \geq \frac{2n}{n-1}(1-2q)^2 \geq 2 v(a).
}
The result follows from the fact that  $v_0(a)<  v(a)$ (see Proposition \ref{prop:fp}).
\ep

Let $\cU\subset [0,1]^n$ be the set of vectors $u$ for which there is  a unique solution $v(u)$ to the fixed-point equation $v=f(u,v)$. The following result  shows the Lipschitz continuity of the function $u\mapsto v(u)$  on $\cU$. 
\begin{proposition}\label{lem:partial}
	For all $u,u'$ in $\cU$, 
	\eqs{
		|v(u) - v(u')| \leq \frac{8}{n} ||u - u'||_1.
	}
\end{proposition}
\bp
	By definition we have $v(u)=f(u,v(u))$ for any $u\in \cU$. Since $\frac{\partial f}{\partial v}>1$ (see Proposition \ref{prop:fps}), by the implicit function theorem, $u\mapsto v(u)$ is differentiable in the interior of $\cU$ and 
	\eqs{
		\forall i=1,\ldots,n,\quad \frac{\partial v}{\partial {u}_i} = \frac{\frac{\partial f}{\partial {u}_i}}{1 - \frac{\partial f}{\partial v}}.
	}
	Observing that 
	$\delta_i(u,v)$ is positive in the interior of $\cU$, we calculate the derivatives of $f$, dropping the arguments $(u,v)$ for convenience:
	\als{	
		\frac{\partial f}{\partial v} &= \frac{1}{(n-2)^2} \left( \sum_{i=1}^n  \sqrt{\delta_i} \right) \left( \sum_{i=1}^n  1/\sqrt{\delta_i} \right),   \sk 
		\frac{\partial f}{\partial {u}_i} &=  - \frac{8(n-1) }{n^2 (n-2)^2} \left( \sum_{j=1}^n  \sqrt{\delta_j/\delta_i} \right). 
	}		
	Now for all $i=1,\ldots,n$,
	\als{
	\frac{\partial f}{\partial v} 
	&= \frac{1}{(n-2)^2} \left[ \left( \sum_{j=1}^n  \sqrt{\delta_j} \right) \left( \sum_{j\neq i}  1/\sqrt{\delta_j} \right) + \sum_{j=1}^n  \sqrt{\delta_j/\delta_i}  \right], \sk
	&\geq \frac{1}{(n-2)^2} \left[ \left( \sum_{j \neq i}  \sqrt{\delta_j} \right) \left( \sum_{j\neq i}  1/\sqrt{\delta_j} \right) +  \sum_{j=1}^n  \sqrt{\delta_j/\delta_i}   \right], \sk
	&\geq \frac{1}{(n-2)^2} \left[ (n-1)^2 + \sum_{j = 1}^n  \sqrt{\delta_j/\delta_i}  \right],\sk
	&\ge 1+ \frac{1}{(n-2)^2}\sum_{j = 1}^n  \sqrt{\delta_j/\delta_i} ,
	}
	where we applied Fact \ref{harm}  to get the second inequality. Thus
	\eqs{
	\frac{\partial f}{\partial v} - 1 \geq \frac{n^2}{8(n-1)} \left|\frac{\partial f}{\partial {u}_i} \right|,
	} 
	and 
	\eqs{
		\left| \frac{\partial v}{\partial {u}_i} \right| \leq \frac{8(n-1)}{n^2}\le \frac{8}{n}.
	}
	Applying the fundamental theorem of calculus yields the result.
\ep

\subsection{Proof}

The proof of Theorem \ref{th:concentration} relies on the following two lemmas, giving concentration inequalities on $\hat a(t)$ and $\hat p(t)$, respectively. 

\begin{lemma}\label{lem:fluctuations}
For any $\epsilon > 0$, we have 
\als{
	  \PP( ||{\hat a}(t) - a||_\infty \geq \epsilon) &\leq 2n\exp\left( - {2 \epsilon^2 \alpha^4 t} \right).
}
\end{lemma}
\bp
 In view of  \eqref{eq:adef}, for all $i=1,\ldots,n$, ${\hat a}_i(t)$ is the sum of $t$ independent, positive random variables bounded by $1/(t\alpha^2)$; in view of \eqref{eq:ri}, we have $\EE[{\hat a}_i(t)]= a_i$. By Hoeffding's inequality,
$$
 \PP( |{\hat a}_i(t) - a_i| \geq \epsilon ) \leq 2\exp\left( - {2 \epsilon^2 \alpha^4 t} \right).
$$
The result follows from the union bound.
\ep

\begin{lemma}\label{lem:final}
Let $\epsilon \in (0,\frac{\gamma}{80}]$. If  $||{\hat a}(t)-a||_\infty \leq \epsilon$ then 
$$||{\hat p}(t)-p||_\infty  \leq  \frac {4  }{\gamma^{3/2}}\epsilon.$$
\end{lemma} 
\bp
Assume that $||{\hat a}(t)-a||_\infty \leq \epsilon$ for some $\epsilon \in (0,\frac{\gamma}{32}]$. Then
\als{
	|v_0({\hat a}(t)) - v_0(a)| &\le  \frac{8 (n-1)}{n^2}||{\hat a}(t)-a||_\infty \leq 8 \epsilon
}
and
\eqs{
| v_1({\hat a}(t)) - v_1(a)| \leq \frac{4}n  ||{\hat a}(t)-a||_1 \leq  4   \epsilon.
}

Since $v_1(a) - v_0(a) > v(a)$ (see Proposition~\ref{pr:lowerboundalpha}) and $v(a)\ge v(a)-v_0(a)=\gamma$, we deduce that
\eqs{
v_1(\hat a(t)) - v_0(\hat a(t)) > \gamma - {12 \epsilon} > 0.
}
By  Proposition~\ref{pr:existancecondition}, the fixed-point equation $v=f(\hat a(t),v)$ has a  unique solution.
By Proposition~\ref{lem:partial},
\begin{equation}\label{eq:ineqv}
	|{v}(\hat a(t)) - v(a)| \leq \frac{8 }{n} ||{\hat a}(t)- a||_1 \leq {8 \epsilon}.
\end{equation}

Now for all $i=1,\ldots,n$,
\begin{align*}
| \hat p_i(t)-p_i|
&= | g_i(\hat a(t),v({\hat a}(t)))-g_i(a,v(a)) |,\\ 
&\leq |g_i(\hat a(t),v(\hat a(t))) - g_i(a,v({\hat a}(t)))|
+ | g_i(a,v({\hat a}(t))) - g_i(a,v(a))|.
\end{align*}
We have
$$
\left|\frac{\partial g_i}{\partial u_i}(u,v)\right| = \frac{n-1}{n \sqrt{\delta_i(u,v)}}\le \frac {1}{\sqrt{\delta_i(u,v)}}  
$$
and
\begin{align*}
\left|\frac{\partial g_i}{\partial v}(u,v)\right|&= \frac{n}{8}\left| \frac{1}{\sqrt{\delta_i(u,v)}} - \frac{1}{\sqrt{v}}\right|\leq \frac n {16}\frac{|\delta_i(u,v)-v|}{\delta_i(u,v)^{3/2}}.
\end{align*}
Since $\delta_i(u,v)\ge v-v_0(u)$, we have
$
\delta_i(a,v(a))\ge \gamma
$
and for any $u$ in the rectangular box formed by $a$ and  $\hat a(t)$, 
$$
\delta_i(u,v(\hat a(t)))\ge v(\hat a(t))-v_0(u) \ge \gamma-16 \epsilon\ge \frac 4 5 \gamma.
$$
Moreover,
$|\delta_i(a,v)-v|\le 4/n$ for any $v$ because $a_i\le 1$ for all $i=1,\ldots,n$, and
$$
\delta_i(a,v)\ge v-v_0(a)\ge \gamma -8  \epsilon \ge \frac  {9\gamma}{10},
$$
for any $v$ between $v(a)$ and $v(\hat a(t))$.
The fundamental theorem of calculus then gives:
$$
	 |g_i(\hat a(t),v(\hat a(t))) - g_i(a,v({\hat a}(t)))|\leq \frac{1 }{ \sqrt{4\gamma/5}}| {\hat a}_i(t)-a_i|
	 $$
	 and
	 $$
	 | g_i(a,v({\hat a}(t))) - g_i(a,v(a))|\leq  \frac 1 {4(9\gamma/10)^{3/2}}{ |{v}(\hat a(t)) - v(a)|}.
$$
We deduce
$$ 
| \hat p_i(t)-p_i|\le   {\sqrt{\frac 5{4 \gamma}}}|{\hat a}_i(t)-a_i| + \frac 14 {\left(\frac {10} {9\gamma}\right)^{3/2}}{ |{v}(\hat a(t)) - v(a)|}.
$$ 
The result then follows from \eqref{eq:ineqv}, on observing that $\gamma  \le 1$ and $\sqrt{5}/2+2(10/9)^{3/2}\le 4$.
\ep

To conclude the proof of Theorem \ref{th:concentration}, we apply Lemmas \ref{lem:fluctuations} and \ref{lem:final}  to obtain
\als{
\PP\left(  ||{\hat p}(t)-p||_\infty   \geq \frac{4}{\gamma^{3/2}} \epsilon \right) &\leq 
						\PP(  ||{\hat a}(t)-a||_\infty \ge \epsilon )
						\leq 2n \exp\left( - {2\epsilon^2 \alpha^4 t}\right),
}
for any $\epsilon \in (0,\frac \gamma {80}]$.
Taking $\varepsilon = \frac{4}{\gamma^{3/2}} \epsilon$ yields the  result, 
on noting  that  $\varepsilon \leq \frac{1}{20}$  and $\gamma \leq 1$ imply $\epsilon \leq \frac{\gamma}{80}$.

\section{Proof of Theorem \lowercase{\ref{th:regret}}}

We control the regret based on the fact that the oracle and our algorithm  output  different answers at time $t$ only if  $W(t) \equiv \frac{1}{n} \sum_{i=1}^n w_i X_i(t)$ and $\frac{1}{n} \sum_{i=1}^n {\hat w}_i(t) X_i(t)$  have different signs. 

We first consider the critical case where $W(t)=0$.
Let $x\in \{-1,0,1\}$ be such that $w^T x = 0$. We have 
\begin{align}
\PP( G(t) = 1 , X(t) = x) &= \PP( G(t) = -1 , X(t) = x)
												= {1 \over 2} \PP( X(t) = x).\label{eq:sym} 
\end{align}
The oracle outputs $G(t)$ with probability ${1 \over 2}$ so that:
\als{
\PP(G^\star (t)\ne G(t) , X(t) = x) &= {1 \over 2} \PP( G(t) = -1 , X(t) = x) + {1 \over 2} \PP( G(t) = 1 , X(t) = x) \sk
	&= {1 \over 2} \PP(X(t) = x)
}   
Now by the independence of  $\hat w(t-1)$ and $X(t)$,
\begin{align*}
\PP( \hat G(t) \neq G(t), X(t) = x)
&=\PP(\hat w(t-1)^T x > 0) \times \PP( G(t) = -1 , X(t) = x)\\
&+\PP(\hat w(t-1)^T x < 0) \times \PP( G(t) = 1 , X(t) = x)  \\
&+\PP(\hat w(t-1)^T x = 0)\times{1 \over 2}\PP( G(t) = 1 , X(t) = x)\\
 & + \PP(\hat w(t-1)^T x = 0)\times{1 \over 2}\PP( G(t) = -1 , X(t) = x).
\end{align*}
In view of \eqref{eq:sym},
$$
\PP( \hat G(t) \neq G(t), X(t) = x) = {1 \over 2} \PP( X(t) = x).
$$
Summing over $x$ such that $w^T x = 0$, we get
$$
\PP(\hat G(t) \ne G(t), W(t)=0)=\PP(G^\star (t)\ne G(t), W(t)=0)
$$
and thus
\als{
\PP(\hat G(t) \ne G(t))-\PP(G^\star (t)\ne G(t))  
= \PP(\hat G(t)\ne G(t) , W(t) \neq 0) -\PP(G^\star (t)\ne G(t), W(t) \neq 0).
}

Now if $W(t)\ne 0$, the oracle and our algorithm will  output  different answers
only if 
\eqs{
\frac{1}{n}\sum_{i=1}^n |{\hat w}_i(t) - w_i| \ge  | W(t) |.
}
Thus we need to bound  the mean estimation error of $w$. 
Assume $ ||{\hat p}(t)-p||_\infty   \leq \frac \eta 2$ and
let $\epsilon_i(t) = \frac 1 \eta {|{\hat p}_i(t)- p_i |}\le 1/2$. 
We have
$$
\hat p_i(t)\ge p_i-\eta \epsilon_i(t)\ge p_i(1-\epsilon_i(t)),
$$
and
$$
1-\hat p_i(t)\le 1-p_i+\eta \epsilon_i(t)\le (1-p_i)(1+\epsilon_i(t)).
$$
We deduce that 
\als{
|{\hat w}_i(t) - w_i | = \left| \log\left(\frac{ p_i (1 - {\hat p}_i(t))}{{\hat p}_i(t)(1- p_i)} \right) \right| 
	\leq \log\left( \frac{1 +\epsilon_i(t)}{1 - \epsilon_i(t)}\right) \leq  {4 \epsilon_i(t)},
}
using inequality $\log z \leq z - 1$ and the fact that $\epsilon_i(t) \leq 1/2$.
Summing the above inequality we get 
\als{
	\frac{1}{n}\sum_{i=1}^n  |{\hat w}_i(t) - w_i| \leq  \frac{4}{n}\sum_{i=1}^n  \epsilon_i(t) 
	= \frac{4}{n \eta}\sum_{i=1}^n  |{\hat p}_i(t)- p_i |
	\le \frac{4 }{ \eta}||\hat p(t)-p||_{\infty}.
}

Now
\begin{align*}
r(t)&= \PP(\hat G(t)\ne G(t))-\PP(G^\star (t)\ne G(t)),\\ 
&=\PP(\hat G(t)\ne G(t), W(t)\ne 0)-\PP(G^\star (t)\ne G(t), W(t)\ne 0),\\ 
&\le  \PP(\hat G(t)\ne G^\star (t),W(t)\ne 0),\\
&\le \PP\left(||\hat p(t)-p||_{\infty} \ge \frac \eta 2\min(|W(t)|/2,1), W(t)\ne 0\right),\\
&\le  \PP\left(||\hat p(t)-p||_{\infty} \ge \frac {\lambda \eta} 4\right).
\end{align*}
The result then follows from Theorem~\ref{th:concentration}.

For the cumulative regret, we use the inequality $\sum_{t \geq 1} e^{-t z} \leq 1/z$, valid for any $z > 0$.

\section{Proof of theorem~\lowercase{\ref{th:concentrationnonstat}}}
Based on the proof for the stationary case, we adopt the following strategy: we first prove that ${\hat a}^{\beta}(t)$ concentrates around $a(t)$ by bounding its bias and fluctuations around its expectation. We then argue that, when ${\hat a}^{\beta}(t)$ is close to $a(t)$ then ${\hat p}(t)$ must be close to $p(t)$. 
\subsection{Preliminary results}

We start by upper bounding the bias of the estimate  $\hat a^\beta (t)$.

\begin{proposition}\label{pr:bias}
We have $|| \EE[{\hat a}^{\beta}(t)] - a(t) ||_{\infty} \leq {2 \sigma \over \beta}$.
\end{proposition}
\bp
	We have:
	\eqs{
	\EE[{\hat a}^{\beta}_i(t)] = \beta \sum_{s=1}^t (1 - \beta)^{t - s} a_i(s).
	}	
	Since 
	$$
	a_i(t)=\frac 1 {n-1} \sum_{j\ne i} (p_i(t)p_j(t)+(1-p_i(t))(1-p_j(t))),
	$$
	we get for all $j \neq i$:
	\als{
	\left|\frac{\partial a_i}{\partial p_i}\right| &= {1 \over n-1} \left|\sum_{j \neq i} (2 p_j - 1) \right| \leq 1,  \sk
	\left|\frac{\partial a_i}{\partial p_j}\right| &= {|2 p_j - 1|\over n-1} \leq {1 \over n-1}.
	}
	We deduce that:
	\als{
		\forall s,t\ge 1,\quad |a_i(s) - a_i(t)| &\leq 2 ||p(s) - p(t)||_{\infty} \leq 2 \sigma |s - t|.
	}
	Hence:
	\als{
	|| \EE[{\hat a}^{\beta}(t)] - a(t) ||_{\infty} &\leq \beta \sum_{s=1}^t (1 - \beta)^{t - s} ||a(s) - a(t)||_{\infty} \sk
	&\leq 2 \sigma \sum_{s=1}^t \beta (1 - \beta)^{t - s} |t -s| \sk
	&\leq 2 \sigma {1 - \beta \over \beta} \leq   {2 \sigma \over \beta}.
	}
\ep

We next provide a concentration inequality for ${\hat a}^{\beta}(t)$.
\begin{proposition}\label{pr:fluctuationabel}
For all $\epsilon \geq 0$,
\eqs{
	\PP( ||{\hat a}^{\beta}(t) - \EE[{\hat a}^{\beta}(t) ] ||_{\infty} \geq \epsilon) \leq 2 n \exp\left( - \frac {2\epsilon^2 \alpha^4}  \beta \right).
	}
\end{proposition}
\bp
 In view of \eqref{eq:abel}, ${\hat a}_i^{\beta}(t)$ is a sum of $t$ positive, independent variables, where the $s$-th variable is bounded by $\beta(1-\beta)^{t-s}\alpha^{-2}$. We have the inequality: 
\eqs{
  \sum_{t \geq 1}  \beta^2 (1-\beta)^{2t} = \frac{\beta}{2 - \beta} \leq \beta.
} 
Hence, from Hoeffding's inequality,
\eqs{
	\PP( |{\hat a}_i^{\beta}(t) - \EE[{\hat a}_i^{\beta}(t) ] | \geq \epsilon) \leq 2 \exp \left( -  \frac {2\epsilon^2 \alpha^4}  \beta \right).
}
The union bound yields the result.	
\ep

\subsection{Proof}
Let $\epsilon \in (0, {\gamma(t) \over 80} - {2 \sigma \over \beta}]$.
Assume that $$||{\hat a}^{\beta}(t) - \EE[{\hat a}^{\beta}(t) ] ||_{\infty} \leq \epsilon.$$
From Proposition~\ref{pr:bias}, this implies
\als{
	||{\hat a}^{\beta}(t) -  a(t) ||_{\infty} &\leq ||{\hat a}^{\beta}(t) -  \EE[ {\hat a}^{\beta}(t)] ||_{\infty} + ||\EE[ {\hat a}^{\beta}(t)] -  a(t) ||_{\infty}, \sk
	&\leq \epsilon + {2 \sigma \over \beta},\sk
	& \le  {\gamma(t) \over 80}.
}
Applying Lemma~\ref{lem:final} yields
\eqs{
||{\hat p}(t)-p(t)||_\infty  \leq  \frac {4}{\gamma^{3 \over 2}(t)} \left( \epsilon + {2 \sigma \over \beta}\right).
}
Applying Proposition~\ref{pr:fluctuationabel} we get the announced result:
\als{
\PP\Big( ||{\hat p}(t)-p(t)||_\infty \geq \frac{4}{\gamma^{3/2}(t)}(\epsilon + \frac {2 \sigma }\beta) \Big) 
   &\leq \PP( ||{\hat a}^{\beta}(t) - \EE[{\hat a}^{\beta}(t) ] ||_{\infty} \leq \epsilon), \sk
   &\leq 2 n \exp\left( - \frac{2 \epsilon^2 \alpha^4 } \beta \right).
}

\end{document}